\documentclass[11pt,letterpaper]{article}
\usepackage{naaclhlt2016}
\usepackage{times}
\usepackage{helvet}
\usepackage{courier}
\usepackage{url}
\usepackage{latexsym}
\usepackage{graphicx}
\usepackage{amsmath}

\naaclfinalcopy 
\def\naaclpaperid{***} 

\title{Combining Recurrent and Convolutional Neural Networks\\ for Relation Classification}

\author{Ngoc Thang Vu$^{1,2}$ \and Heike Adel${^1}$ \and Pankaj Gupta${^3}$ \and Hinrich Sch\"{u}tze$^{1}$ \\
       $^1$Center for Information and Language Processing, LMU Munich\\
       Oettingenstr. 67, 80538 Munich, Germany\\
       $^2$Institute for Natural Language Processing, University of Stuttgart\\
       Pfaffenwaldring 5b, 70569 Stuttgart, Germany\\
       $^3$Siemens Corporate Technology - Knowledge Modeling and Retrieval\\
       Otto-Hahn-Ring 6, 81739 Munich, Germany\\
       {\tt thang.vu@ims.uni-stuttgart.de | heike.adel@cis.lmu.de} \\{\tt gupta.pankaj.ext@siemens.com | inquiries@cislmu.org}}

\date{}

\def\eqref#1{Equation~\ref{eqn:#1}}
\def\eqlabel#1{\label{eqn:#1}}

\def\eqref#1{Equation~\ref{eqn:#1}}
\def\eqlabel#1{\label{eqn:#1}}

\begin{document}
\maketitle
\begin{abstract}
This paper investigates two different neural architectures for the task of relation classification:
convolutional neural networks and recurrent neural networks.
For both models, we demonstrate the effect of different architectural choices.
We present a new context representation for convolutional neural networks for relation 
classification (extended middle context).
Furthermore, we propose connectionist bi-directional recurrent neural networks
and introduce ranking loss for their optimization.
Finally, we show that combining convolutional and recurrent neural networks using 
a simple voting scheme is accurate enough to improve results. 
Our neural models achieve state-of-the-art results on
the SemEval 2010 relation classification task.
\end{abstract}

\section{Introduction}
Relation classification is the task of assigning sentences with two marked entities to a predefined
set of relations.
The sentence ``We poured the \texttt{<e1>}milk\texttt{</e1>} 
into the \texttt{<e2>}pumpkin mixture\texttt{</e2>}.'', 
for example, expresses the relation \texttt{Entity-Destination(e1,e2)}.
While early research mostly focused on support vector machines or maximum entropy classifiers \cite{svmSystem,meSystem}, 
recent research showed performance improvements
by applying neural networks (NNs) \cite{socher,zeng2014,yu2014,nguyen,deSantos2015,zhang} on the benchmark data
from Sem\-Eval 2010 shared task 8~\cite{semevalData} .

This study investigates two different types of NNs: 
recurrent neural networks (RNNs) and convolutional neural networks (CNNs) 
as well as their combination. We make the following contributions:

(1) We propose  \emph{extended middle context},
a new context representation for CNNs for relation classification.
The extended middle context uses all parts of the sentence (the relation arguments, left of the relation arguments,
between the arguments, right of the arguments) and pays special attention to the middle part.

(2) We present \emph{connectionist bi-directional RNN
  models} which are especially suited for sentence
classification tasks since they combine all intermediate
hidden layers for their final decision. Furthermore, the
ranking loss function is introduced for the RNN model
optimization which has not been  investigated in the
literature for relation classification before.

(3) Finally, we combine CNNs and RNNs using a simple voting scheme 
and achieve new state-of-the-art results on the SemEval 2010 benchmark dataset.

\section{Related Work}
\label{relatedWork}
In 2010, manually annotated data for relation classification was released 
in the context of a SemEval shared task~\cite{semevalData}.
Shared task participants used, i.a., support vector machines 
or maximum entropy classifiers \cite{svmSystem,meSystem}.
Recently, their results on this data set were outperformed
by applying NNs \cite{socher,zeng2014,yu2014,nguyen,deSantos2015}.

\newcite{zeng2014} built a CNN based only on the context between the relation arguments and extended it with several lexical features.
\newcite{kim2014} and others used convolutional filters of different sizes for CNNs.
\newcite{nguyen} applied this to relation classification and obtained improvements over single filter sizes.
\newcite{deSantos2015} replaced the softmax layer of the CNN with a ranking layer. 
They showed improvements and published the best result so
far on the SemEval dataset, to our knowledge.

\newcite{socher} used another NN architecture for relation classification: 
recursive neural networks that built
recursive sentence representations based on syntactic parsing.
In contrast, \newcite{zhang}
investigated a temporal structured RNN with only words as input. They used a bi-directional model
with a pooling layer on top.

\section{Convolutional Neural Networks (CNN)}
\label{CNNs}
CNNs perform a discrete convolution on an input matrix
with a set of different filters. For NLP tasks, the input matrix represents a sentence:
Each column of the matrix stores the word embedding of the corresponding word.
By applying a filter with a width of, e.g., three columns, three neighboring words (trigram) are
convolved. 
Afterwards, the results of the convolution are pooled. Following \newcite{collobertWeston}, we
perform max-pooling which extracts the maximum value for each filter
and, thus, the most informative n-gram for
the following steps. 
Finally, the resulting values are concatenated and used for classifying the relation expressed in the sentence.

\subsection{Input: Extended Middle Context}

\begin{figure}
\centering
\includegraphics[width=\columnwidth]{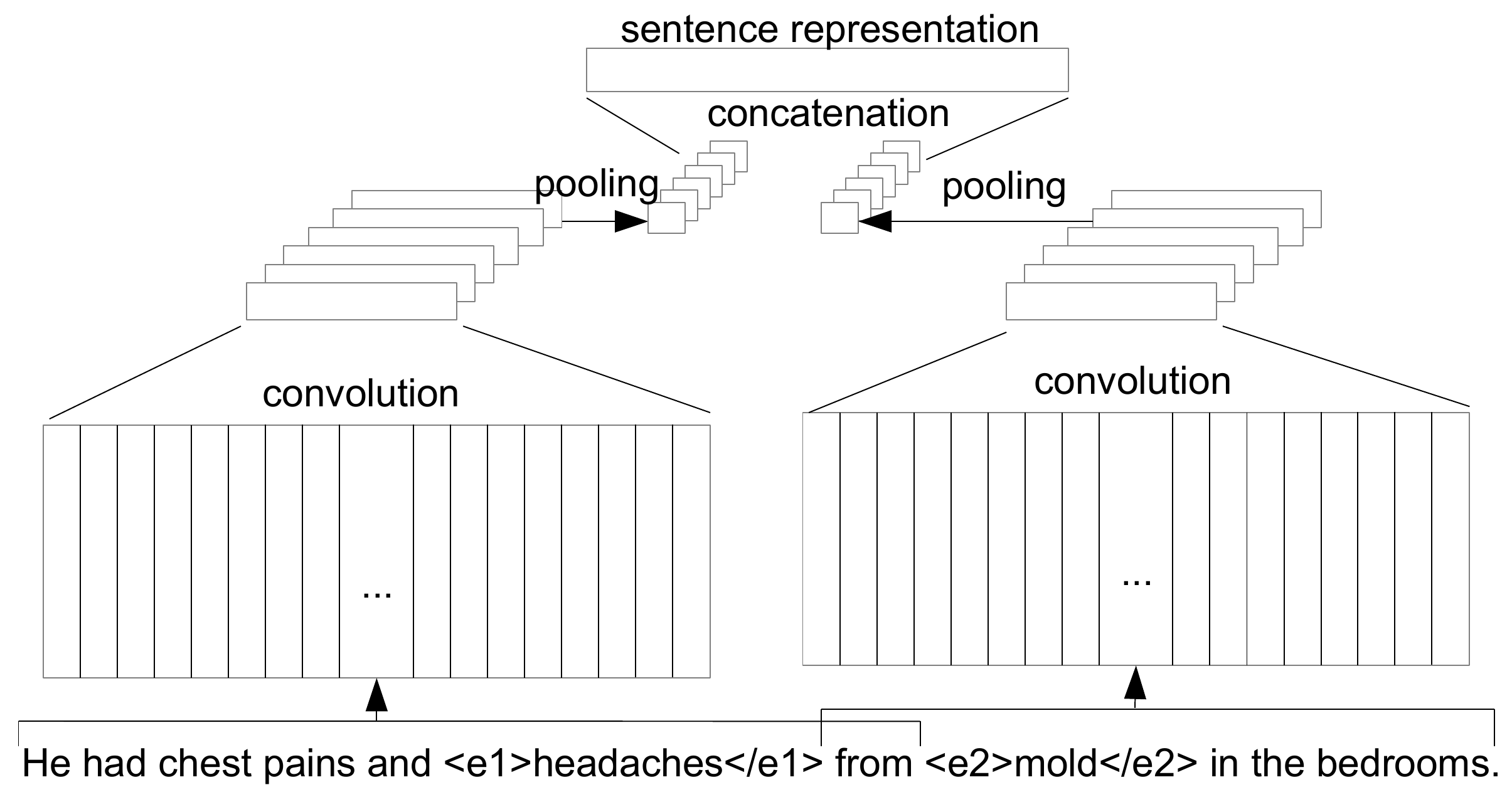}
\caption{CNN with extended contexts}
\label{extendedMiddleContext}
\end{figure}

One of our contributions is a new input representation especially
designed for relation classification. The contexts are split into three disjoint regions based on the two relation arguments: the left  context, the middle context and the 
right context. Since in most cases the middle context contains the most relevant information for the relation, we
want to focus on it but not ignore the other regions completely. Hence, we propose to
use two contexts: (1) a combination of the left context, the left entity and the middle context; and (2) a 
combination of the middle context, the right entity and the right context. 
Due to the repetition of the middle context, we force the network to
pay special attention to it. The two contexts are processed by two independent convolutional and max-pooling layers.
After pooling, the results are concatenated to form the sentence representation.
Figure~\ref{extendedMiddleContext} depicts this procedure. It shows an examplary sentence: 
``He had chest pain and \texttt{<e1>}headaches\texttt{</e1>} from \texttt{<e2>}mold\texttt{</e2>} in the bedroom.''
If we only considered the middle context ``from'', the network might be tempted to predict a relation like \texttt{Entity-Origin(e1,e2)}.
However, by also taking the left and right context into account, the model can detect the relation \texttt{Cause-Effect(e2,e1)}.
While this could also be achieved by integrating the whole context into the model, using the whole context
can have disadvantages for
longer sentences: The max pooling step can easily choose a value from a part of the sentence which is
far away from the mention of the relation. With splitting the context into two parts, we reduce this danger.
Repeating the middle context increases the chance for the max pooling step to pick a value from the
middle context.

\subsection{Convolutional Layer}
Following previous work (e.g., \cite{nguyen}, \cite{deSantos2015}), 
we use 2D filters spanning all embedding dimensions.
After convolution, a max pooling operation is applied that stores only
the highest activation of each filter.
We apply filters with different window 
sizes 2-5 (multi-windows) as in \cite{nguyen}, 
i.e. spanning a different number of input words. 

\section{Recurrent Neural Networks (RNN)}
Traditional RNNs consist of an input vector, a history vector and an output vector.
Based on the representation of the current input word and the previous
history vector, a new history is computed. Then, an output is predicted (e.g., using a softmax layer).
In contrast to most traditional RNN architectures, we use the RNN for sentence modeling,
i.e., we predict an output vector only after processing the whole sentence
and not after each word.
Training is performed using backpropagation through time \cite{bptt} which
unfolds the recurrent computations of the history vector for a certain number of time steps.
To avoid exploding gradients, we use gradient clipping with a threshold of 10~\cite{gradientClipping}.

\subsection{Input of the RNNs}
Initial experiments showed that using trigrams as input instead of single words
led to superior results. Hence, at timestep $t$ we do not only give word $w_t$
to the model but the trigram $w_{t-1} w_t w_{t+1}$ by concatenating the corresponding
word embeddings.

\subsection{Connectionist Bi-directional RNNs}
\begin{figure}
\includegraphics[width=\columnwidth]{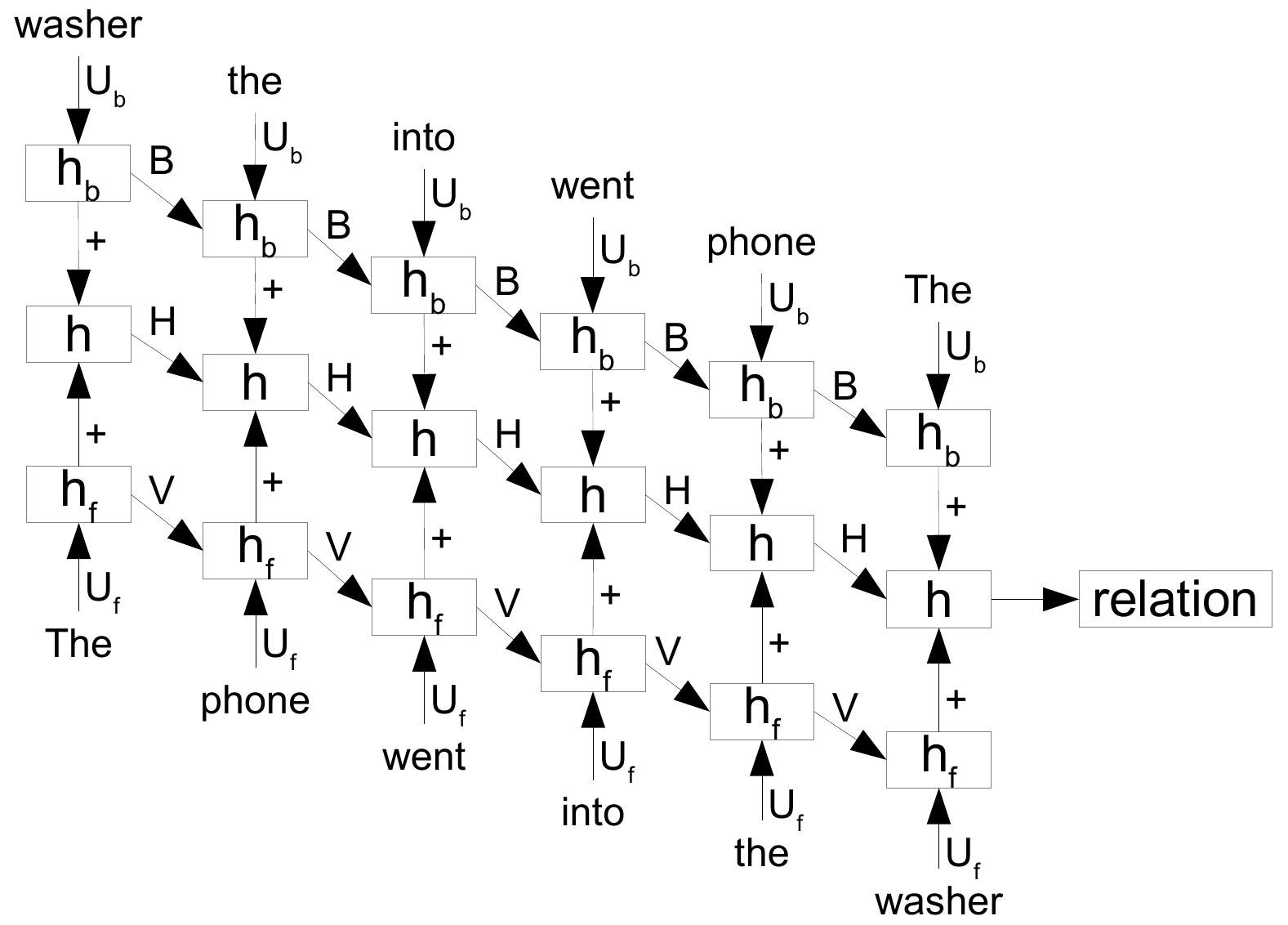}
\caption{Connectionist bi-directional RNN}
\label{bidirectionalRNN}
\end{figure}

Especially for relation classification, the processing of the relation arguments 
might be easier with knowledge of the succeeding words.
Therefore in bi-directional RNNs, not only a
history vector of word $w_t$ is regarded but also a future vector.
This leads to the following conditioned probability
for the history $h_t$ at time step $t \in [1,n]$:
\begin{eqnarray}
h_{f_t} = f(U_f \cdot w_t + V \cdot h_{f_{t-1}}) \label{eq:ft}\\
h_{b_t} = f(U_b \cdot w_{n-t+1} + B \cdot h_{b_{t+1}}) \label{eq:bt}\\
h_t = f(h_{b_t} + h_{f_t} + H \cdot h_{t-1}) \label{eq:connectionist}
\end{eqnarray}
Thus, the network can be split into three parts: a
forward pass which processes the original sentence word by word (Equation \ref{eq:ft});
a backward pass which processes the reversed sentence
word by word (Equation \ref{eq:bt}); and a combination of both (Equation \ref{eq:connectionist}).
All three parts are trained jointly.
This is also depicted in Figure \ref{bidirectionalRNN}.

Combining forward and backward pass by
adding their hidden layer is similar to~\cite{zhang}.
We, however, also add a connection to the 
previous combined hidden layer with weight $H$ to
be able to include all intermediate hidden layers 
into the final decision of the network (see Equation 
\ref{eq:connectionist}). We call this
``connectionist bi-directional RNN''.

In our experiments, we compare this RNN
with uni-directional RNNs and bi-directional RNNs without
additional hidden layer connections.

\section{Model Training}

\subsection{Word Representations}
Words are represented by concatenated vectors: a word embedding and a position feature vector.

\noindent \textbf{Pretrained word embeddings.}
In this study, we used the word2vec toolkit \cite{word2vec}
to train embeddings on an English Wikipedia from May 2014.
We only considered words appearing more than 100 times 
and added a special PADDING token
for convolution. This results in an embedding training text
of about 485,000 terms and $6.7 \cdot 10^9$ tokens.
During model training, the embeddings are updated. 

\noindent \textbf{Position features.}
We incorporate randomly initialized position embeddings
similar to 
\newcite{zeng2014}, \newcite{nguyen}
and \newcite{deSantos2015}.
In our RNN experiments, we investigate different possibilities of integrating position
information: position embeddings, position embeddings with entity presence flags
(flags indicating whether the current word is one of the relation arguments),
and position indicators \cite{zhang}. 

\subsection{Objective Function: Ranking Loss}

\noindent \textbf{Ranking.}
We applied the ranking loss function proposed in \newcite{deSantos2015} to train our models. 
It maximizes the distance between 
the true label $y^+$ and the best competitive label $c^-$ 
given a data point $x$. The objective function is
\begin{equation}\eqlabel{obj}
\begin{split}
  L = \log(1+\exp(\gamma(m^+ - s_{\theta}(x)_{y^+}))) \\ 
     + \log(1+\exp(\gamma(m^- + s_{\theta}(x)_{c^-})))
\end{split}
\end{equation}
with $s_{\theta}(x)_{y^+}$ and $s_{\theta}(x)_{c^-}$ being the scores for the classes $y^+$ and $c^-$ respectively.
The parameter $\gamma$ controls the penalization of the prediction errors and $m^+$ and $m^-$ are margins
for the correct and incorrect classes.
Following \newcite{deSantos2015}, we set $\gamma = 2, m^+ = 2.5, m^- = 0.5$.
We do not learn a pattern for the class \texttt{Other} 
but increase its difference to the best competitive label by
using only the second summand in \eqref{obj} during training.

\section{Experiments and Results}
\label{experiments}
We used the relation classification dataset of the Sem\-Eval 2010 task 8~\cite{semevalData}.
It consists of sentences which have been manually labeled with 19 relations 
(9 directed relations and one artificial 
class \texttt{Other}).
8,000 sentences have been distributed as training set and 2,717
sentences served as test set.
For evaluation, we applied the official scoring script and report the
macro F1 score which also served as the official result of the shared task.

RNN and CNN models were implemented with theano~\cite{theano1,theano2}.
For all our models, we use L2 regularization with a weight of 0.0001.
For CNN training, we use mini batches of 25 training examples while
we perform stochastic gradient descent for the RNN. 
The initial learning rates are 0.2 for the CNN and 0.01 for the RNN.
We train the models for 10 (CNN) and 50 (RNN) epochs without
early stopping. As activation function, we apply tanh for the CNN
and capped ReLU for the RNN.
For tuning the hyperparameters, we split the training data
into two parts: 6.5k (training) and 1.5k (development) sentences.
We also tuned the learning rate schedule on dev.

Beside of training single models, we also report ensemble
results for which we combined the presented single 
models with a voting process.

\subsection{Performance of CNNs}
\label{architecture}
As a baseline system, we implemented a CNN similar to the
one described by \newcite{zeng2014}. It consists of a
standard convolutional layer with filters with only one window
size, followed by a softmax layer.  As input it
uses the middle context. In contrast to \newcite{zeng2014}, 
our CNN does not have an additional fully connected hidden layer.
Therefore, we
increased the number of convolutional filters to 1200 to keep
the number of parameters comparable.
With this, we obtain a baseline result of 73.0.  After
including 5 dimensional position features, the performance was improved to
78.6 (comparable to 78.9 as reported by \newcite{zeng2014}
without linguistic features).

In the next step, we investigate how this result changes if we successively 
add further features to our CNN: multi-windows
for convolution (window sizes: 2,3,4,5 and 300 feature maps each), ranking 
layer instead of softmax 
and our proposed extended middle context. 
Table~\ref{architectureExperiments} shows the results.
Note that all numbers are produced by CNNs with a comparable number of parameters.
We also report F1 for increasing the word embedding 
dimensionality from 50 to 400. 
The position embedding dimensionality is 5 in combination with 50 dimensional
word embeddings and 35 with 400 dimensional word embeddings.
Our results show that especially the ranking layer 
and the embedding size have an important impact on the performance.

\begin{table}
\centering
\small
\begin{tabular}{l|l}
CNN & F1\\
\hline
Baseline (emb dim: 50) & 73.0 \\
+ position features & 78.6* \\
+ multi-windows features map & 78.7 \\
+ ranking layer & 81.9*\\
+ extended middle context & 82.2\\
\hline
+ increase emb dim to 400 & 83.9*\\
\hline
ensemble & \textbf{84.2}
\end{tabular}
\caption{F1 score of CNN and its components, * indicates statisticial significance
compared to the result in the line above (z-test, $p < 0.05$)}
\label{architectureExperiments}
\end{table}

\subsection{Performance of RNNs}
\begin{table}
\centering
\small
\begin{tabular}{l|l}
RNN & F1\\
\hline
uni-directional (Baseline, emb dim: 50) & 61.2\\
uni-directional  + position embs & 68.3*\\
uni-directional  + position embs + entity flag & 73.1*\\
uni-directional  + position indicators & 73.4\\
\hline
bi-directional + position indicators & 74.2*\\
\hline
connectionist-bi-directional+position indicators & 78.4*\\
+ ranking layer &
81.4*\\
+ increase emb dim to 400 & 
82.5*\\
\hline
ensemble & \textbf{83.4}
\end{tabular}
\caption{F1 score of RNN and its components, * indicates statisticial significance
compared to the result in the line above (z-test, $p < 0.05$)}
\label{resultsRNN}
\end{table}
As a baseline for the RNN models, we apply a uni-directional
RNN which predicts the relation after processing the whole sentence.
With this model, we achieve an
F1 score of 61.2 on the SemEval test set.

Afterwards, we investigate the impact of different position features
on the performance of uni-directional RNNs (position embeddings, 
position embeddings concatenated with a flag indicating
whether the current word is an entity or not, and position indicators \cite{zhang}).
The results indicate that position indicators (i.e.
artificial words that indicate the entity presence)
perform the best on the SemEval data. We achieve an F1 score
of 73.4 with them. However, the difference to using position embeddings with
entity flags is not statistically significant.

Similar to our CNN experiments, we successively vary the RNN models
by using bi-directionality, by adding connections between the hidden layers
(``connectionist''), by applying ranking instead of softmax to predict the
relation and by increasing the word embedding dimension to 400.

The results in Table \ref{resultsRNN} show that 
all of these variations lead to statistically significant improvements.
Especially the additional hidden layer connections and 
the integration of the ranking layer
have a large impact on the performance.

\subsection{Combination of CNNs and RNNs}
Finally, we combine our CNN and RNN models using a voting process.
For each sentence in the test set, we apply several CNN and RNN models
presented in Tables~\ref{architectureExperiments} and \ref{resultsRNN}
and predict the class with the most votes. In case of
a tie, we pick one of the most frequent classes randomly.
The combination achieves an F1 score of 84.9 which is better than the performance of the two 
NN types alone. 
It, thus, confirms our assumption that the networks provide complementary information: 
while the RNN computes a weighted combination of all words in the sentence,
the CNN extracts the most informative n-grams for the relation 
and only considers their resulting activations.

\begin{table}
\centering
\small
\begin{tabular}{l|lc}
Classifier & F1 \\
\hline
SVM \cite{rink2010utd} & 82.2 \\ 
\hline
RNN \cite{socher} & 77.6 \\
MVRNN \cite{socher} & 82.4\\
\hline
CNN  \cite{zeng2014} & 82.7 \\
\hline
FCM \cite{yu2014} & 83.0 \\
\hline
bi-RNN \cite{zhang} & 82.5\\
\hline
CR-CNN \cite{deSantos2015} & 84.1\\
\hline
\hline
R-RNN & 83.4\\
ER-CNN & 84.2\\
ER-CNN + R-RNN & \textbf{84.9}\\
\end{tabular}
\caption{State-of-the-art results for relation classification}
\label{stateOfTheArt}
\end{table}

\subsection{Comparison with State of the Art}
Table \ref{stateOfTheArt} shows the results of our models 
ER-CNN (extended ranking CNN) and R-RNN (ranking RNN)
in the context of other state-of-the-art models.
Our proposed models obtain state-of-the-art results on the 
Sem\-Eval 2010 task 8 data set without making use of 
any linguistic features. 

\section{Conclusion}
\label{conclusion}
In this paper, we investigated different features and architectural 
choices for convolutional and recurrent neural networks 
for relation classification without using any linguistic features.
For convolutional neural networks, we presented a new 
context representation for relation classification. 
Furthermore, we introduced connectionist 
recurrent neural networks for sentence classification tasks and
performed the first experiments with ranking recurrent neural networks.
Finally, we showed that even a simple combination of convolutional and 
recurrent neural networks improved results.
With our neural models, we achieved new state-of-the-art results on the SemEval 2010
task 8 benchmark data.

\section*{Acknowledgments}
Heike Adel is a recipient of the Google European Doctoral
Fellowship in Natural Language Processing and this
research is supported by this fellowship.

This research was also supported by Deutsche
Forschungsgemeinschaft: grant SCHU 2246/4-2.

\bibliographystyle{naaclhlt2016}

\bibliography{refs}

\end{document}